\documentstyle[11pt,epsbox]{article}

\title{\bf Selective Sampling of Effective Example Sentence Sets for
  Word Sense Disambiguation} \author{\normalsize\bf FUJII Atsushi,
  INUI Kentaro, TOKUNAGA Takenobu and TANAKA Hozumi} \date{Department
  of Computer Science
  \\ Tokyo Institute of Technology \\ 2-12-1 Oookayama Meguroku Tokyo
  152, JAPAN \\ \smallskip {\tt
    \{fujii,inui,take,tanaka\}@cs.titech.ac.jp}}

\makeatletter
\@definecounter{subfloatnumber}
\def\@T@bLe{table}
\def\subfloatcap{-\alph{subfloatnumber}} 

\newif\if@subfloatcaptionhead \@subfloatcaptionheadtrue
\def\NoSubfloatCaptionHead{\@subfloatcaptionheadfalse}

\newif\if@subfloatwithcaptionhead
\def\subcaption{\@ifstar{%
 \@subfloatwithcaptionheadfalse\subc@ption}{%
 \@subfloatwithcaptionheadtrue\subc@ption}}

\def\subc@ption{\@addtoreset{subfloatnumber}{\@captype}%
\ifx\@captype\@T@bLe\relax\else\addtocounter{\@captype}{1}\fi%
\def\the@subfloatnumber{\csname the\@captype\endcsname\subfloatcap}%
\stepcounter{subfloatnumber}%
\edef\@currentlabel{\csname the@subfloatnumber\endcsname}%
\@dblarg{\@subcaption\@captype}}

\long\def\@subcaption#1[#2]#3{\par\addcontentsline{\csname
  ext@#1\endcsname}{#1}{\protect\numberline{\csname
  the#1\endcsname\subfloatcap}{\ignorespaces #2}}\begingroup
    \@parboxrestore
    \small
   \if@subfloatcaptionhead
    \if@subfloatwithcaptionhead
     \@makecaption{\csname fnum@#1\endcsname\subfloatcap}%
         {\ignorespaces #3}\par
    \else
     \@makecaptionWOheading{\subfloatcap}{\ignorespaces #3}
    \fi
   \else\@makecaptionWOheading{}{\ignorespaces #3}\par\fi
    \endgroup%
     \ifx\@captype\@T@bLe\relax\else\addtocounter{\@captype}{-1}\fi}

\long\def\@makecaptionWOheading#1#2{%
   \vskip 10\p@
   \setbox\@tempboxa\hbox{#1 #2}%
   \ifdim \wd\@tempboxa >\hsize
       #1 #2\par
     \else
       \hbox to\hsize{\hfil\box\@tempboxa\hfil}%
   \fi}
\makeatother

\begin{document}

\maketitle\thispagestyle{empty}

\newcommand{\E}{\mbox{$e$}}
\newcommand{\EXi}[2]{\mbox{${\cal E}_{#1,#2}$}}
\newcommand{\V}{\mbox{$v$}}
\newcommand{\SS}{\mbox{$s$}}
\newcommand{\SSi}[1]{\mbox{$s_{#1}$}}
\newcommand{\C}{\mbox{$c$}}
\newcommand{\Ci}[1]{\mbox{$c_{#1}$}}
\newcommand{\Ni}[1]{\mbox{$n_{#1}$}}
\newcommand{\Mi}[1]{\mbox{$m_{#1}$}}
\newcommand{\eq}[1]{(\ref{#1})}
\newcommand{\set}[1]{\mbox{\bf #1}}

\begin{abstract}
  This paper proposes an efficient example selection method for
  example-based word sense disambiguation systems. To construct a
  practical size database, a considerable overhead for manual sense
  disambiguation is required.  Our method is characterized by the
  reliance on the notion of the training utility: the degree to which
  each example is informative for future example selection when used for
  the training of the system. The system progressively collects examples
  by selecting those with greatest utility.  The paper reports the
  effectivity of our method through experiments on about one thousand
  sentences. Compared to experiments with random example selection, our
  method reduced the overhead without the degeneration of the
  performance of the system.
\end{abstract}

\section{Introduction}
\label{sec:intro}

Word sense disambiguation is a crucial task in many NLP applications,
such as machine translation~\cite{brown:91}, parsing~\cite
{lytinen:86,k.nagao:94} and text
retrieval~\cite{krovets:92,voorhees:93}.  Given the growing
utilization of machine readable texts, word sense disambiguation
techniques have been variously used in corpus-based
approaches~\cite{brown:91,dagan:94,fujii:96:d,kurohashi:94,niwa:94,schutze:92:a,uramoto:94,yarowsky:95}.
Unlike rule-based approaches, corpus-based approaches release us from
the task of generalizing observed phenomena in order to disambiguate
word senses. Our system is based on such an approach, or more
precisely it is based on an example-based approach~\cite{fujii:96:d}. 
Since this approach requires a certain number of examples of
disambiguated verbs, we have to carry out this task manually, that is,
we disambiguate verbs appearing in a corpus prior to their use by the
system. A preliminary experiment on ten Japanese verbs showed that the
system needed on average about one hundred examples for each verb in
order to achieve 82\% of accuracy in disambiguating verb senses.  In
order to build an operational system, the following problems have to
be taken into account:
\begin{enumerate}
\item Since there are about one thousand basic verbs in Japanese, a
  considerable overhead is associated with manual word sense
  disambiguation.
\item Given human resource limitations, it is not reasonable to
  manually analyze large corpora as they can provide virtually
  infinite input.
\item Given the fact that example-based natural language systems,
  including our system, search the example-database (database,
  hereafter) for the most similar examples with regard to the input,
  the computational cost becomes prohibitive if one works with a very
  large database size~\cite{kudo:95}.
\end{enumerate}
All these problems suggest a different approach, namely to {\it
  select\/} a small number of optimally informative examples from a
given corpora. Hereafter we will call these examples ``samples.''

\begin{figure}[htbp]
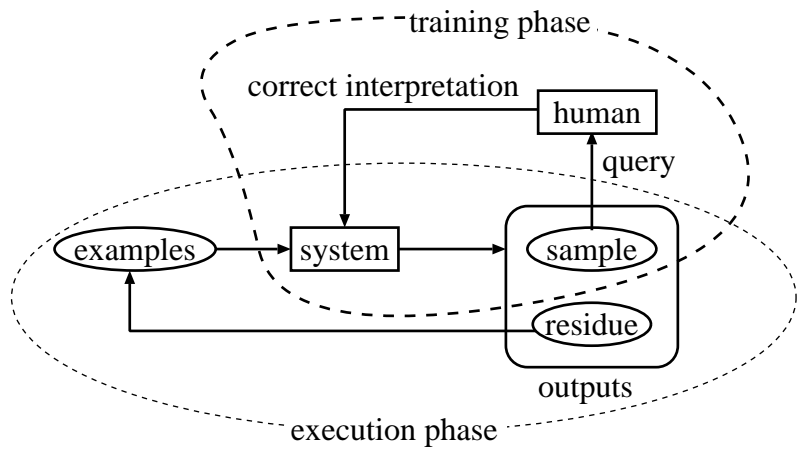

  \begin{center}
    \leavevmode \epsfile{hscale=0.9,vscale=0.9,file=concept.eps}
  \end{center}
  \caption{Flow of control of the example sampling system}
  \label{fig:concept}
\end{figure}

Our method, based on the utility maximization principle, decides on
which examples should be included in the database.  This decision
procedure is usually called {\it selective sampling}. Selective
sampling directly addresses the first two problems mentioned above. 
The overall control flow of systems based on selective sampling can be
depicted as in figure \ref{fig:concept}, where ``system'' refers to
dedicated NLP applications. The sampling process basically cycles
between the execution and the training phases. During the execution
phase, the system generates an interpretation for each example, in
terms of parts-of-speech, text categories or word senses.  During the
training phase, the system selects samples for training from the
previously produced outputs. During this phase, a human expert
provides the correct interpretation of the samples so that the system
can then be trained for the execution of the remaining data. Several
researchers have proposed such an approach.

Lewis et al. proposed an example sampling method for statistics-based
text classification~\cite{lewis:94}. In this method, the system always
selects samples which are not certain with respect to the correctness
of the answer. Dagan et al. proposed a committee-based sampling
method, which is currently applied to HMM training for part-of-speech
tagging~\cite{dagan:95:b}. This method selects samples based on the
training utility factor of the examples, i.e. the informativity of the
data with respect to future training. However, as all these methods
are implemented for statistics-based models, there is a need to
explore how to formalize and map these concepts into the example-based
approach.

With respect to problem 3, a possible solution would be the
generalization of redundant examples~\cite{kaji:92,nomiyama:93}.
However, such an approach implies a significant overhead for the
manual training of each example prior to the generalization. This
shortcoming is precisely what our approach allows to avoid: reducing
both the overhead as well as the size of the database.

Section \ref{sec:vader} briefly describes our method for a verb sense
disambiguation system. The next Section \ref{sec:sampling} elaborates
on the example sampling method, while section \ref{sec:eval} reports
on the results of our experiment.  Before concluding in section
\ref{sec:conclusion}, discussion is added in section
\ref{sec:discussion}.

\section{Example-based verb sense disambiguation system}
\label{sec:vader}

\begin{figure}[htbp]
  \begin{center}
    \small
    \leavevmode
    \begin{tabular}{|lll|} \hline
      & & \\
      \framebox{{\it toru\/}:} & & \\ & & \\ \hline
      $\left\{\begin{tabular}{ll} {\it suri\/} & (pickpocket) \\
      {\it kanojo\/} & (she) \\ {\it ani\/} & (brother)
    \end{tabular}\right\}$ {\it ga\/} &
      $\left\{\begin{tabular}{ll} {\it kane\/} & (money) \\
      {\it saifu\/} & (wallet) \\ {\it otoko\/} & (man) \\
    {\it uma\/} & (horse) \\ {\it aidea\/} & (idea)
  \end{tabular}\right\}$ {\it o\/} &
    {\it toru\/} (to take/steal) \\ \hline
      $\left\{\begin{tabular}{ll} {\it kare\/} & (he) \\
      {\it kanojo\/} & (she) \\
      {\it gakusei\/} & (student)
    \end{tabular}\right\}$ {\it ga\/} &
      $\left\{\begin{tabular}{ll} {\it menkyosh\^{o}\/} & (license) \\
      {\it shikaku\/} & (qualification) \\
      {\it biza\/} & (visa)
  \end{tabular}\right\}$ {\it o\/} &
    {\it toru\/} (to attain) \\ \hline
      $\left\{\begin{tabular}{ll} {\it kare\/} & (he) \\
      {\it chichi\/} & (father) \\ {\it kyaku\/} & (client)
    \end{tabular}\right\}$ {\it ga\/} &
      $\left\{\begin{tabular}{ll} {\it shinbun\/} & (newspaper) \\
      {\it zasshi\/} & (journal)
  \end{tabular}\right\}$ {\it o\/} &
    {\it toru\/} (to subscribe) \\ \hline
      $\left\{\begin{tabular}{ll} {\it kare\/} & (he) \\
      {\it dantai\/} & (group) \\
      {\it ryok\^{o}kyaku\/} & (passenger) \\
      {\it joshu\/} & (assistant)
    \end{tabular}\right\}$ {\it ga\/} &
      $\left\{\begin{tabular}{ll} {\it kippu\/} & (ticket) \\
      {\it heya\/} & (room) \\ {\it hik\^{o}ki\/} & (airplane)
  \end{tabular}\right\}$ {\it o\/} &
    {\it toru\/} (to reserve) \\ \hline
    {\hfill \centering $\vdots$ \hfill} & {\hfill \centering $\vdots$
      \hfill} & {\hfill \centering $\vdots$ \hfill} \\ \hline
\end{tabular}
  \end{center}
  \caption{A fragment of a database, and the entry
    associated with the Japanese verb {\it toru\/}}
  \label{fig:toru}
\end{figure}

\begin{figure}[htbp]
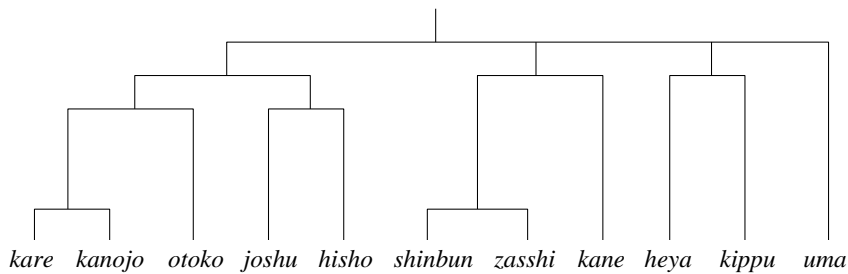

  \begin{center}
    \leavevmode \epsfile{hscale=0.7,vscale=0.7,file=bgh.eps}
  \end{center}
  \caption{A fragment of {\it Bunruigoihyo\/}}
  \label{fig:bgh}
\end{figure}

Our method for disambiguating verb senses uses a database containing
examples of collocations for each verb sense and its associated case
frame(s). Figure \ref{fig:toru} shows a fragment of the entry
associated with the Japanese verb {\it toru}. As with most words, the
verb {\it toru\/} has multiple senses, a sample of which are ``to
take/steal,'' ``to attain,'' ``to subscribe'' and ``to reserve.'' The
database specifies the case frame(s) associated with each verb sense. 
In Japanese, a complement of a verb consists of a noun phrase (case
filler) and its case marker suffix, for example {\it ga\/}
(nominative) or {\it o\/} (accusative). The database lists several
case filler examples for each case. The task of the system is ``to
interpret'' the verbs occurring in the input text, i.e. to choose one
sense from among a set of candidates.  All verb senses we use are
defined in ``IPAL''~\cite{ipal:87}, a machine readable dictionary. 
IPAL also contains example case fillers as shown in figure
\ref{fig:toru}. Given an input, in our case a simple sentence, the
system identifies the verb sense on the basis of the scored similarity
between the input and the examples given for each verb sense.  Let us
take as an example the sentence below:
\begin{list}{}{\setlength{\leftmargin}{0mm}}
\item
  \begin{tabular}{ccc}
    {\it hisho\/} {\it ga\/} & {\it shindaisha\/} {\it o\/} & {\it
      toru}. \\ (secretary-NOM) & (sleeping car-ACC) & (?)
  \end{tabular}
\end{list}
In this example, one may consider {\it hisho\/} (``secretary'') and
{\it shindaisha\/} (``sleeping car'') to be semantically similar to
{\it joshu\/} (``assistant'') and {\it hik\^{o}ki\/} (``airplane'')
respectively, and since both collocate with the ``to reserve'' sense
of {\it toru\/} one could infer that {\it toru\/} may be interpreted
as ``to reserve.'' The similarity between two different case fillers
is estimated according to the length of the path between them in a
thesaurus. Our current experiments are based around the Japanese word
thesaurus {\it Bunruigoihyo\/}~\cite {bgh:64}. Figure \ref{fig:bgh}
shows a fragment of {\it Bunruigoihyo\/} including some of the nouns
in both figure \ref{fig:toru} and the example sentence above, with
each word corresponding to a leaf in the structure of the thesaurus.
As with most thesauri, the length of the path between two terms in
{\it Bunruigoihyo} is expected to reflect their relative similarity. 
In table \ref{tab:kuro}, we show our measure of similarity, based on
the length of the path between two terms, as proposed by Kurohashi et
al~\cite{kurohashi:94}.

\begin{figure}[htbp]
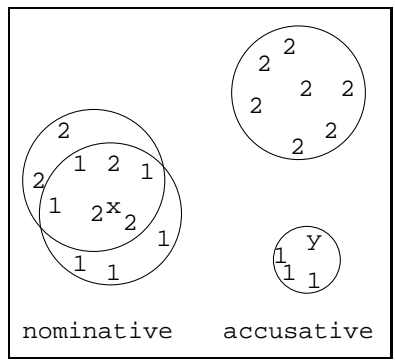

  \begin{center}
    \leavevmode
    \fbox{\parbox{7cm}{
        \centering
        \smallskip
        \epsfile{hscale=0.7,vscale=0.7,file=ccd.eps}
        }}
  \end{center}
  \caption{The semantic ranges of the nominative and accusative with
    verb {\it toru\/}}
  \label{fig:ccd}
\end{figure}

Furthermore, since the restrictions imposed by the case fillers in
choosing the verb sense are not equally selective, we consider a
weighted case contribution to the disambiguation (CCD) of the verb
senses.  This CCD factor is taken into account when computing the
score of a verb's sense. Consider again the case of {\it toru} in
figure \ref{fig:toru}. Since the semantic range of nouns collocating
with the verb in the nominative does not seem to have a strong
delinearization in a semantic sense (in figure \ref{fig:toru}, the
nominative of each verb sense displays the same general concept, i.e.
animate), it would be difficult, or even risky, to properly interpret
the verb sense based on the similarity in the nominative. In contrast,
since the ranges are diverse in the accusative, it would be feasible
to rely more strongly on the similarity here.  This argument can be
illustrated as in figure \ref{fig:ccd}, in which the symbols ``{\tt
  1}'' and ``{\tt 2}'' denote example case fillers of different case
frames, and an input sentence includes two case fillers denoted by
``x'' and ``y.''

The figure shows the distribution of example case fillers for the
respective case frames, denoted in a semantic space.  The semantic
similarity between two given case fillers is represented by the
physical distance between two symbols.  In the nominative, since
``{\tt x}'' happens to be much closer to a ``{\tt 2}'' than any ``{\tt
  1},'' ``{\tt x}'' may be estimated to belong to the range of ``{\tt
  2}''s, although ``{\tt x}'' actually belongs to both sets of ``{\tt
  1}''s and ``{\tt 2}''s.  In the accusative, however, ``{\tt y}''
would be properly estimated to belong to the set of ``{\tt 1}''s due
to the mutual independence of the two accusative case filler sets,
even though examples did not fully cover each of the ranges of ``{\tt
  1}''s and ``{\tt 2}''s. Note that this difference would be critical
if example data were sparse.  We will explain the method used to
compute CCD later in this section.

To illustrate the overall algorithm, we will consider an abstract
specification of both input and the database (see figure
\ref{fig:case-frame}). Let the input be \{$\Ni{\Ci{1}}$-$\Mi{\Ci{1}}$,
$\Ni{\Ci{2}}$-$\Mi{\Ci{2}}$, $\Ni{\Ci{3}}$-$\Mi{\Ci{3}}$, $\V$\},
where $\Ni{\Ci{i}}$ denotes the case filler for the case $\Ci{i}$, and
$\Mi{\Ci{i}}$ denotes the case marker for $\Ci {i}$. The
interpretation candidates for $\V$ are derived from the database as
$\SSi{1}$, $\SSi{2}$ and $\SSi{3}$. The database contains also a set
$\EXi{\SSi{i}}{\Ci{j}}$ of case filler examples for each case $\Ci{j}$
of each sense $\SSi{i}$ (``---'' indicates that the corresponding case
is not allowed).

\begin{table}[htbp]
  \caption{The relation between the length of the path between two nouns
    $X$ and $Y$ ($len(X,Y)$) in {\it Bunruigoihyo\/} and their relative
    similarity ($sim(X,Y)$)}
  \label{tab:kuro}
  \centering
  \medskip
  \begin{tabular}{|c|ccccccc|} \hline
    $len(X,Y)$ & 0 & 2 & 4 & 6 & 8 & 10 & 12 \\ \hline $sim(X,Y)$ & 11 &
    10 & 9 & 8 & 7 & 5 & 0 \\ \hline
  \end{tabular}
\end{table}
  
\begin{figure}[htbp]
  \centering
  \begin{tabular}{|c|lllll|} \hline
    input & $\Ni{\Ci{1}}$-$\Mi{\Ci{1}}$ & $\Ni{\Ci{2}}$-$\Mi{\Ci{2}}$ &
    $\Ni{\Ci{3}}$-$\Mi{\Ci{3}}$ & & $\V$ (?) \\ \hline \hline
    & $\EXi{\SSi{1}}{\Ci{1}}$ &
    $\EXi{\SSi{1}}{\Ci{2}}$ & $\EXi{\SSi{1}}{\Ci{3}}$ &
    --- & $\V$ ($\SSi{1}$) \\
    database & $\EXi{\SSi{2}}{\Ci{1}}$ &
    $\EXi{\SSi{2}}{\Ci{2}}$ & $\EXi{\SSi{2}}{\Ci{3}}$ &
    $\EXi{\SSi{2}}{\Ci{4}}$ & $\V$ ($\SSi{2}$) \\
    & --- &
    $\EXi{\SSi{3}}{\Ci{2}}$ & $\EXi{\SSi{3}}{\Ci{3}}$ &
    --- & $\V$ ($\SSi{3}$) \\ \hline
  \end{tabular}
  \caption{An input and the database}
  \label{fig:case-frame}
\end{figure}

During the verb sense disambiguation process, the system discards
first those candidates whose case frame does not fit the input. In the
case of figure \ref{fig:case-frame}, $\SSi{3}$ is discarded because
the case frame of $\V$ ($\SSi{3}$) does not subcategorize for the case
$\Ci{1}$.

In the next step the system computes the score of the remaining
candidates and chooses as the most plausible interpretation the one
with the highest score. The score of an interpretation is computed by
considering the {\em weighted\/} average of the similarity degrees of
the input complements with respect to each of the example case fillers
(in the corresponding case) listed in the database for the sense under
evaluation. Formally, this is expressed by equation \eq{eq:score},
where $S(\SS)$ is the score of the sense $\SS$ of the input verb, and
$SIM(\Ni{\C},\EXi{\SS}{\C})$ is the maximum similarity degree between
the input complement $\Ni{\C}$ and the corresponding complements in
the database example $\EXi{\SS}{\C}$ (equation \eq{eq:sim}).
\begin{equation}
\label{eq:score}
S(\SS) = \frac{\textstyle \sum_{\C} SIM(\Ni{\C},\EXi{\SS}{\C})\cdot
  CCD(\C)}{\textstyle \sum_{\C} CCD(\C)}
\end{equation}
\begin{equation}
  \label{eq:sim}
  SIM(\Ni{\C},\EXi{\SS}{\C}) = {\displaystyle \max_{\E \in
      \EXi{\SS}{\C}}} sim(\Ni{\C},\E)
\end{equation}
In equation \eq{eq:sim}, $sim$ stands for the similarity degree
between $\Ni{\C}$ and an example case filler $\E$ as given by table
\ref{tab:kuro}.

$CCD(\C)$ expresses the weight factor of the case $c$ contribution to
the (current) verb sense disambiguation.  Intuitively preference
should be given to cases displaying case fillers which are classified
in semantic categories of greater independence. Let $\V$ be a verb
with $n$ senses ($\SSi{1}, \SSi{2}, \ldots, \SSi{n}$) and let
$\EXi{\SSi{i}}{\C}$ be the set of example case fillers for the case
$\C$, associated with the sense $\SSi{i}$. Then, $\C$'s contribution
to $\V$'s sense disambiguation, $CCD(\C)$, is likely to be higher if
the example case filler sets \{$\EXi{\SSi{i}}{\C}~|~i = 1, \ldots,
n$\} share less elements.  The notion of sharing is defined based on
the similarity as in equation \eq{eq:x}.
\begin{equation}
  \label{eq:x}
  \{X\} \cup \{Y\} = \{X\}~~\mbox{\ if\ } sim(X,Y) >= 9
\end{equation}
With these definitions, $CCD(\C)$ is given by equation \eq{eq:ccd}.
\begin{equation}
  \begin{array}{l}
    CCD(\C) = {\displaystyle \left(\frac {\displaystyle
        1}{\displaystyle _{\it n\/}C_{2}}\sum_{{\it i\/} = 1}^{{\it
          n\/}-1}\sum_{{\it j\/} = {\it i\/} + 1}^{{\it
          n\/}}\frac{\displaystyle
        |\EXi{\SSi{i}}{\C}|+|\EXi{\SSi{j}}{\C}|-2|\EXi{\SSi{i}}{\C}\cap
        \EXi {\SSi{j}}{\C}|}{\displaystyle
        |\EXi{\SSi{i}}{\C}|+|\EXi{\SSi{j}}{\C}|}\right)^\alpha
      \label{eq:ccd}}
  \end{array}
\end{equation}
Where $\alpha$ is the constant for parameterizing the extent to which
CCD influences verb sense disambiguation.  The larger $\alpha$, the
stronger CCD's influence on the system's output.

\section{Example sampling algorithm}
\label{sec:sampling}

\subsection{Overview}
\label{subsec:overview}

Let us look again at figure \ref{fig:concept} in section \ref
{sec:intro}. In this diagram, ``outputs'' refers to a corpus in which
each sentence is assigned the proper interpretation of the verb during
the execution phase. In the ``training'' phase, the system stores
samples of manually disambiguated verb senses (simply checked or
appropriately corrected by a human) in the database to be later used
in a new execution phase. This is the issue we turn to in this
section.

Lewis et al. proposed the notion of uncertain example sampling for the
training of statistics-based text classifiers~\cite{lewis:94}. Their
method selects those examples that the system classifies (in this
case, matching a text category) with minimum certainty. This method is
based on the assumption that there is no need for teaching the system
the correct answer when it answered with high certainty. However, we
should take into account the training effect a given example has on
other examples. In other words, by selecting an appropriate example as
a sample, we can get more correct examples in the next cycle of
iteration. In consequence, the number of examples to be taught will
decrease. We consider maximization of this effect by means of a
training utility function (TUF) aiming at ensuring that the example
with the highest training utility figure, is the most useful example
at a given point in time.

Let $\set{S}$ be a set of sentences, i.e. a given corpus, and
$\set{T}$ be a subset of $\set{S}$ in which each sentence has already
been manually disambiguated for training. In other words, sentences in
$\set{T}$ have been selected as samples, and are hence stored in the
database.  Let $\set{X}$ be the set of the residue, realizing equation
\eq {eq:corpus}.
\begin{equation}
  \label{eq:corpus}
  \set{S} = \set{X} \cup \set{T}
\end{equation}

We introduce a utility function $TUF(x)$, which computes the training
utility figure for an example $x$. The sampling algorithm gives
preference to examples of maximum utility, by way of equation
\eq{eq:argmax}.
\begin{equation}
  \label{eq:argmax}
  \arg \max_{x \in \set{X}} TUF(x)
\end{equation}

We will explain in the following sections how one could estimate TUF,
based on the estimation of the certainty figure of an interpretation.
Ideally the sampling size, i.e. the number of samples selected at each
iteration would be such as to avoid retraining of similar examples. It
should be noted that this can be a critical problem for
statistics-based
approaches~\cite{brown:91,dagan:94,niwa:94,schutze:92:a,yarowsky:95},
as the reconstruction of statistic classifiers is expensive. However,
example-based systems~\cite{fujii:96:d,kurohashi:94,uramoto:94} do not
require the reconstruction of the system, but examples have to be
stored in the database.  It also should be noted that in each
iteration, the system needs only compute the similarity between each
example $x$ belonging to $\set{X}$ and the newly stored example,
instead of every example belonging to $\set{T}$, because of the
following reasons:
\begin{itemize}
\item storing an example of verb sense interpretation
  $\SSi{i}$, will not affect the score of other verb senses,
\item if the system memorizes the current score of $\SSi{i}$ for each
  $x$, the system simply needs to compare it with the newly computed
  score between $x$ and the newly stored example in $\set{T}$ and
  choose the greater of the two to be the new plausibility of
  $\SSi{i}$.
\end{itemize}
This reduces the time complexity of each iteration from $O(N^2)$ to
$O(N)$, given that $N$ is the total number of examples in $\set{S}$.

\subsection{Interpretation certainty}
\label{subsec:certainty}

Lewis et al. estimate certainty of an interpretation by the ratio
between the probability of the most plausible text category, and the
probability of any other text category, excluding the most probable
one.  Similarly, in our example-based verb sense disambiguation
system, we introduce the notion of interpretation certainty of
examples based on the following applicability restrictions:
\begin{enumerate}
\item the highest interpretation score is sufficiently large,
\item the highest interpretation score is significantly larger than the
  second highest score.
\end{enumerate}
The rationale for these restrictions is given below. Consider figure
\ref{fig:certainty}, where each symbol denotes an example in
$\set{S}$, with symbols ``{\tt x}'' belonging to $\set{X}$ and symbols
``{\tt e}'' belonging to $\set{T}$. The curved lines delimit the
semantic vicinities (extents) of the two ``{\tt e}''s, i.e. sense 1
and sense 2, respectively\footnote{Note that this method can easily be
  extended for a verb which has more than two senses. In section
  \ref{sec:eval}, we conducted an experiment using multiply ambiguous
  verbs.}. The semantic similarity between two sentences is
graphically portrayed by the physical distance between the two symbols
representing them. In figure \ref{fig:certainty-a}, ``{\tt x}''s
located inside a semantic vicinity are expected to be interpreted with
high certainty as being similar to the appropriate example ``{\tt
  e},'' a fact which is in line with restriction 1 mentioned above. 
However, in figure \ref{fig:certainty-b}, the degree of certainty for
the interpretation of any ``{\tt x}'' which is located inside the
intersection of the two semantic vicinities cannot be great.  This
happens when the case fillers of two or more verb senses are not
selective enough to allow a clear cut delineation among them. This
situation is explicitly rejected by restriction 2.

\begin{figure}[htbp]
  \begin{center}
    \leavevmode
    \begin{minipage}[t]{.47\textwidth}
      \centering
      \epsfile{hscale=0.7,vscale=0.7,file=certainty-a.eps}
      \subcaption{The case where the interpretation certainty of the
        enclosed ``{\tt x}'' is great}
      \label{fig:certainty-a}
    \end{minipage}
    \hfill
    \begin{minipage}[t]{.47\textwidth}
      \centering
      \epsfile{hscale=0.7,vscale=0.7,file=certainty-b.eps}
      \subcaption{The case where the interpretation certainty of the
        the enclosed ``{\tt x}'' is small}
      \label{fig:certainty-b} \medskip
    \end{minipage}
  \end{center}
  \caption{The concept of interpretation certainty}
  \label{fig:certainty}
\end{figure}

Considering the two restrictions, we compute interpretation
certainties by using equation \eq{eq:certainty}, where $C(x)$ is the
interpretation certainty of an example $x$. $S_1(x)$ and $S_2(x)$ are
the highest and second highest scores for $x$, respectively. 
$\lambda$, which ranges from 0 to 1, is a parametric constant to
control the degree to which each condition affects the computation of
$C(x)$.
\begin{equation}
  \label{eq:certainty}
  C(x) = \lambda\cdot S_1(x) + (1 - \lambda)\cdot(S_1(x) - S_2(x))
\end{equation}

We estimated the validity of the notion of the interpretation
certainty through a preliminary experiment, in which we used the same
corpus used for another experiment as described in section \ref
{sec:eval}.  In this experiment, we conducted a six-fold cross
validation, that is, we divided the training/test data into six equal
parts, and conducted six trials in which a different part was used as
test data each time, and the rest as training data. We shall call
these two sets the ``test set'' and the ``training set.''  Thereafter,
we evaluated the relation between the applicability and the precision
of the system.

\begin{figure}[htbp]
  \begin{center}
    \leavevmode \epsfile{hscale=0.6,vscale=0.6,file=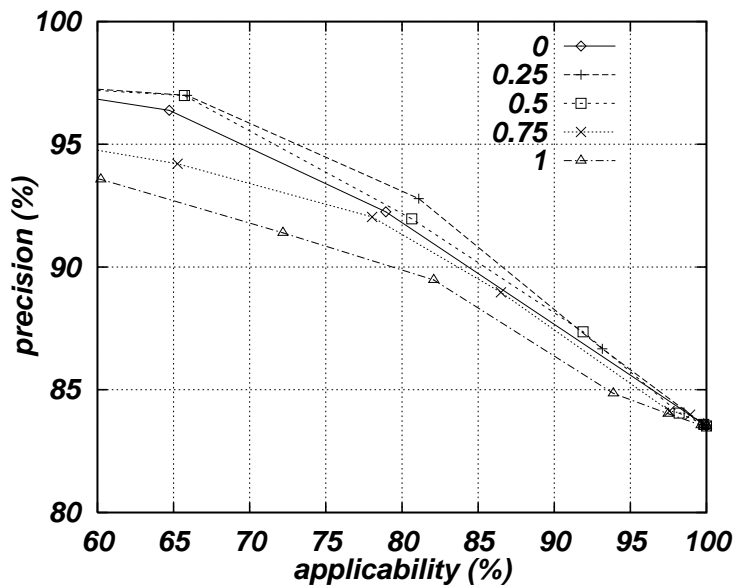}
  \end{center}
  \caption{The relation between applicability and precision with several
    $\lambda$'s} \label{fig:app-pre}
\end{figure}

In this experiment, the applicability is the ratio between the number
of cases where the certainty of the system's interpretation of the
outputs is above a certain threshold, and the number of inputs. The
precision is the ratio between the number of correct outputs, and the
number of inputs. Increasing the value of the threshold, the precision
also increases (at least theoretically), while the applicability
decreases. Figure \ref{fig:app-pre} shows the result of the experiment
with several values of $\lambda$, in which the optimal $\lambda$ value
seems to be in the range 0.25 to 0.5. It can be seen that, as we
assumed, both restrictions are essential for the estimation of the
interpretation certainty.

\subsection{Training utility}
\label{subsec:utility}

\begin{figure}[htbp]
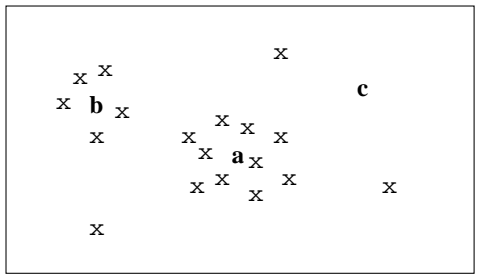

  \begin{center}
    \leavevmode
    \epsfile{hscale=0.7,vscale=0.7,file=utility.eps}
  \end{center}
  \caption{The concept of training utility} \label{fig:utility}
\end{figure}

The training utility of an example ``{\tt a}'' is greater than that of
another example ``{\tt b}'' when the total interpretation certainty of
examples in $\set{X}$ increases more after training using the example
``{\tt a}'' than after using the example ``{\tt b}.'' Let us consider
figure \ref{fig:utility}, with the basic notation as in figure
\ref{fig:certainty}, and let us compare the training utility of the
examples ``{\tt a},'' ``{\tt b}'' and ``{\tt c}.'' Note that in this
figure, whatever example we use for training, the interpretation
certainty for the neighbours (``{\tt x}''s) of the chosen example
increases. However, it is obvious that we can increase the total
interpretation certainty of ``{\tt x}''s when we use ``{\tt a}'' for
training as it has more neighbours than either ``{\tt b}'' or ``{\tt
  c}.'' In consequence, one can expect that the size of the database,
which is directly proportional to the number of training examples, can
be decreased. Let $\Delta C(x\!=\!\SS,y)$ be the difference in the
interpretation certainty of $y\in\set{X}$ after training with
$x\in\set{X}$ taken with the sense $\SS$. $TUF(x\!=\!s)$, which is the
training utility function for $x$ taken with sense $\SS$, can be
computed by equation \eq{eq:utility}.
\begin{equation}
  \label{eq:utility}
  TUF(x\!=\!\SS) = \sum_{y \in \set{X}}\Delta C(x\!=\!\SS,y)
\end{equation}
We compute $TUF(x)$ by calculating the average of each
$TUF(x\!=\!\SS)$, weighted by the probability that $x$ takes sense
$\SS$. This can be realized by equation \eq{eq:utility_sum}, where
$P(x\!=\! \SS)$ is the probability that $x$ is used in training with
the sense $\SS$.
\begin{equation}
  \label{eq:utility_sum}
  TUF(x) = \sum_{\SS}P(x\!=\!\SS)\cdot TUF(x\!=\!\SS)
\end{equation}
Given the fact that (a) $P(x\!=\!\SS)$ is difficult to estimate in the
current formulation, and (b) the cost of computation for each
$TUF(x\!=\!\SS)$ is not trivial, we temporarily approximate $TUF(x)$
as in equation \eq{eq:utility_temp}, where $\set{K}$ is a set of the
$k$-best verb sense(s) of $x$ with respect to the interpretation score
in the current state.
\begin{equation}
  \label{eq:utility_temp}
  TUF(x) \simeq \sum_{s \in \set{K}}\frac{\textstyle 1}{\textstyle
    k}\cdot TUF(x\!=\!s)
\end{equation}

\section{Evaluation}
\label{sec:eval}

We compared the performance of our example sampling method with random
sampling, in which a certain proportion of a given corpus is randomly
selected for training. We compared the two sampling methods by
evaluating the relation between various numbers of examples in
training, and the performance of the system on another corpus.  We
conducted a six-fold cross validation as described in section
\ref{subsec:certainty}, but in this experiment, each method selected
some proportion of the training set as samples.  We used the same
corpus as described in table \ref{tab:corpus} as training/test data.
Both sampling methods used examples from IPAL to initialize the system
(as seeds) with the number of example case fillers for each case being
on average of about 3.7.

The training/test data used in the experiment contained about one
thousand simple Japanese sentences collected from news articles.  Each
of the sentences in the training/test data used in our experiment
contained one or several complement(s) followed by one of the ten
verbs enumerated in table \ref{tab:corpus}. In table \ref{tab:corpus},
the column of ``English gloss'' describes typical English translations
of the Japanese verbs.  The column of ``\# of sentences'' denotes the
number of sentences in the corpus, ``\# of senses'' denotes the number
of verb senses based on IPAL, and ``lower bound'' denotes the
precision gained by using a naive method, where the system
systematically chooses the most frequently appearing interpretation in
the training data~\cite{gale:92}.
\begin{table}[htbp]
  \caption{The corpus used for the experiments}
  \begin{center}
    \leavevmode
    \medskip
    \begin{tabular}{|c||c|c|c|c|} \hline
      verb & English gloss & \# of sentences & \# of senses & lower
      bound \\ \hline \hline
      {\it ataeru\/} & give & 136 & 4 & 66.9 \\ \hline
      {\it kakeru\/} & hang & 160 & 29 & 25.6 \\ \hline
      {\it kuwaeru\/} & add & 167 & 5 & 53.9 \\ \hline
      {\it noru\/} & ride & 126 & 10 & 45.2 \\ \hline
      {\it osameru\/} & govern & 108 & 8 & 25.0 \\ \hline
      {\it tsukuru\/} & make & 126 & 15 & 19.8 \\ \hline
      {\it toru\/} & take & 84 & 29 & 26.2 \\ \hline
      {\it umu\/} & bear offspring & 90 & 2 & 81.1 \\ \hline
      {\it wakaru\/} & understand & 60 & 5 & 48.3 \\ \hline
      {\it yameru\/} & stop & 54 & 2 & 59.3 \\ \hline \hline
      total & --- & 1111 & --- & 43.7 \\ \hline
    \end{tabular}
  \end{center}
  \label{tab:corpus}
\end{table}

We at first estimated the system's performance by its precision, that
is the ratio of the number of correct outputs, compared to the number
of inputs.  In this experiment, we set $\lambda = 0.5$ in equation
\eq{eq:certainty}, and $k = 1$ in equation \eq{eq:utility_temp}.  The
influence of CCD, i.e. $\alpha$ in equation \eq{eq:ccd}, was extremely
large so that the system virtually relied solely on the SIM of the
case with the greatest CCD.

\begin{figure}[htbp]
  \begin{center}
    \leavevmode
    \epsfile{hscale=0.6,vscale=0.6,file=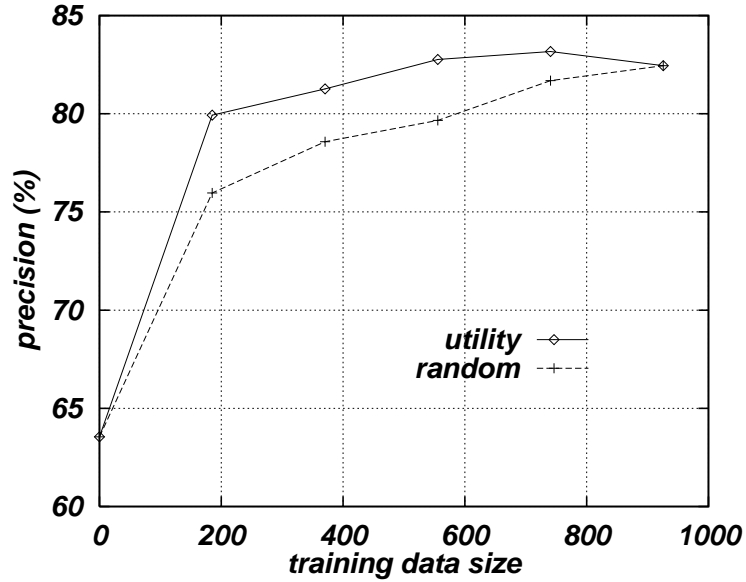}
  \end{center}
  \caption{The relation between the training data size and precision
    of the system}
  \label{fig:precision}
\end{figure}

Figure \ref{fig:precision} shows the relation between the size of the
training data and the precision of the system. In figure \ref
{fig:precision}, when the x-axis is zero, the system has used only the
seeds given by IPAL. It should be noted that with the final step,
where all examples in the training set have been provided to the
database, the precision of both methods is equal.  Looking at figure
\ref{fig:precision} one can see that the precision of random sampling
was surpassed by our training utility sampling method. It solves the
first two problems mentioned in section \ref{sec:intro}. One can also
see that the size of the database can be reduced without degrading the
system's precision, and as such it can solve the third problem
mentioned in section \ref{sec:intro}.

We further evaluated the system's performance in the following way.
Integrated with other NLP systems, the task of our verb sense
disambiguation system is not only to output the most plausible verb
sense, but also the interpretation certainty of its output, so that
other systems can vary the degree of reliance on our system's output. 
The following are properties which are required for our system:
\begin{itemize}
\item the system should output as many correct answers as possible,
\item the system should output correct answers with great
  interpretation certainty,
\item the system should output incorrect answers with diminished
  interpretation certainty.
\end{itemize}
Motivated by these properties, we formulated a new performance
estimation measure, PM, as shown in equation \eq{eq:performance}.  A
greater accuracy of performance of the system will lead to a greater
PM value.
\begin{equation}
  \label{eq:performance}
  PM = \frac{\textstyle 1}{\textstyle
    N}\sum_{x}\delta\cdot\frac{\textstyle C(x)}{\textstyle C_{max}}
\end{equation}
In equation \eq{eq:performance}, $C_{max}$ is the maximum value of the
interpretation certainty, which can be derived by substituting the
maximum and the minimum interpretation score for $S_1(x)$ and
$S_2(x)$, respectively, in equation \eq{eq:certainty}.  Following
table \ref{tab:kuro}, we assign 11 and 0 to be the maximum and the
minimum of the interpretation score, and therefore $C_{max}$ = 11,
disregarding the value of $\lambda$ in equation \eq{eq:certainty}. $N$
is the total number of the inputs and $\delta$ is a coefficient
defined as in equation \eq{eq:delta}.
\begin{equation}
  \label{eq:delta}
  \delta = \left\{
  \begin{array}{lr}
    1 & \mbox{if the interpretation of $x$ is correct} \\
    \noalign{\vskip 1ex}
    -p & \mbox{otherwise}
  \end{array}
\right.
\end{equation}
In equation \eq{eq:delta}, $p$ is the parametric constant to control
the degree of the penalty for a system error.  For our experiment, we
set $p=1$, meaning that PM was in the range $-1$ to 1.

\begin{figure}[htbp]
  \begin{center}
    \leavevmode
    \epsfile{hscale=0.6,vscale=0.6,file=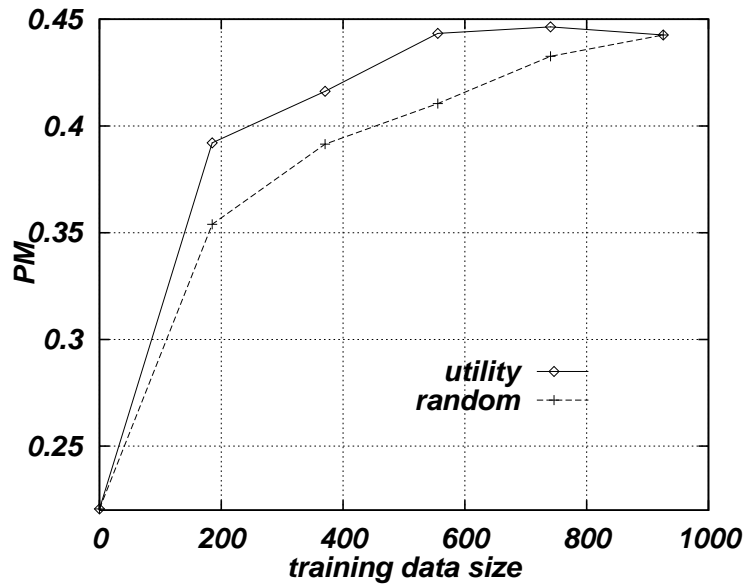}
  \end{center}
  \caption{The relation between the training data size and performance
    of the system}
  \label{fig:performance}
\end{figure}

Figure \ref{fig:performance} shows the relation between the size of
the training data and the value of PM. In this experiment, it can be
seen that the performance of random sampling was again surpassed by
our training utility sampling method, and the size of the database can
be reduced without degrading the system's performance.

\section{Discussion}
\label{sec:discussion}

In this section, we will discuss several remaining problems.  First,
since in equation \eq{eq:utility}, the system calculates the
similarity between $x$ and each example in $\set{X}$, computation of
$TUF(x\!=\!s)$ becomes time consuming.  To avoid this problem, a method used
in efficient database search techniques~\cite{kolodner:93,utsuro:94},
in which the system can search some neighbour examples of $x$ with
optimal time complexity, can be potentially used.

Second, there is a problem as to when to stop the training: that is,
as mentioned in section \ref{sec:intro}, it is not reasonable to
manually analyze large corpora as they can provide virtually infinite
input.  One plausible solution would be to select a point when the
increment of the total interpretation certainty of remaining examples
in $\set{X}$ is not expected to exceed a certain threshold.

Finally, we should also take the semantic ambiguity of case fillers
(noun) into account. Let us consider figure \ref{fig:uncertain}, where
the basic notation is the same as in figure \ref{fig:certainty}, and
one possible problem caused by case filler ambiguity is illustrated.
Let ``{\tt x1}'' and ``{\tt x2}'' denote different senses of a case
filler ``{\tt x}.''  Following the basis of equation
\eq{eq:certainty}, the interpretation certainty of ``{\tt x}'' is
small in both figure \ref{fig:uncertain-a} and \ref{fig:uncertain-b}.
However, in the situation as in figure \ref{fig:uncertain-b}, since
(a) the task of distinction between the {\it verb\/} senses 1 and 2 is
easier, and (b) instances where the sense ambiguity of case fillers
corresponds to distinct verb senses will be rare, training using
either ``{\tt x1}'' or ``{\tt x2}'' will be less effective than as in
figure \ref{fig:uncertain-a}.  It should also be noted that since {\it
  Bunruigoihyo\/} is a relatively small-sized thesaurus and does not
enumerate many word senses, this problem is not critical in our case.
However, given other existing thesauri like the EDR electronic
dictionary~\cite{edr:93} or WordNet~\cite{wordnet:93}, these two
situations should be strictly differentiated.

\begin{figure}[htbp]
  \begin{center}
    \leavevmode
    \begin{minipage}[t]{.47\textwidth}
      \centering
      \epsfile{hscale=0.7,vscale=0.7,file=uncertain-a.eps}
      \subcaption{Interpretation certainty of ``{\tt x}'' is small because
        ``{\tt x}'' lies in the intersection of distinct verb senses}
      \label{fig:uncertain-a}
    \end{minipage}
    \hfill
    \begin{minipage}[t]{.47\textwidth}
      \centering
      \epsfile{hscale=0.7,vscale=0.7,file=uncertain-b.eps}
      \subcaption{Interpretation certainty of ``{\tt x}'' is small because
        ``{\tt x}'' is semantically ambiguous}
      \label{fig:uncertain-b} \medskip
    \end{minipage}
  \end{center}
  \caption{Two separate sceneries where the interpretation certainty
    of ``{\tt x}'' is small}
  \label{fig:uncertain}
\end{figure}

\section{Conclusion}
\label{sec:conclusion}

In this paper we proposed an example sampling method for example-based
verb sense disambiguation. We also reported on the system's
performance by way of experiments. The experiments showed that our
method, which is based on the notion of training utility, has reduced
the overhead for the training of the system, as well as the size of
the database.

As pointed out in section \ref{sec:intro}, the generalization of
examples~\cite{kaji:92,nomiyama:93} is another method for reducing the
size of the database. Whether coupling these two methods would
increase overall effectivity is an empirical matter requiring further
exploration.

Future work will include more sophisticated methods for verb sense
disambiguation and methods of acquiring seeds, the acquisition of
which is currently based on an existing dictionary.  We will also
build an experimental database for natural language processing using
our example sampling method.

\section*{Acknowledgments}

The authors would like to thank Dr. Manabu Okumura (JAIST, Japan), Mr.
Timothy Baldwin (TITech, Japan), and Dr. Michael Zock and Dr. Dan
Tufis (LIMSI, France) for their comments on an earlier version of this
paper.

\bibliographystyle{plain}

\end{document}